\documentclass{article} % For LaTeX2e
\usepackage{iclr2027_conference,times,graphicx,booktabs,multirow,float,xcolor}

\newcommand{\avgnum}[2]{\makebox[2.6em][r]{#1}\makebox[3.3em][l]{\,#2}}

\usepackage{amsmath,amsfonts,bm}

\def\eqref#1{equation~\ref{#1}}
\def\1{\bm{1}}

\DeclareMathAlphabet{\mathsfit}{\encodingdefault}{\sfdefault}{m}{sl}
\SetMathAlphabet{\mathsfit}{bold}{\encodingdefault}{\sfdefault}{bx}{n}

\newcommand{\E}{\mathbb{E}}

\newcommand{\KL}{D_{\mathrm{KL}}}

\usepackage{hyperref}
\hypersetup{hidelinks}
\usepackage{url}

\title{Instruct, Not Answer: Using Instruction Privileges in On-Policy Context Distillation}

\author{\parbox[t]{\dimexpr\textwidth-2\tabcolsep\relax}{\centering\normalfont \textbf{Hantao Yu$^{1}$ \thanks{Work done during an internship at Amazon Web Services.} \hspace{0.1em}, Xiaoxue Han$^{2}$, Udaya Ghai$^{2}$, Ferhat Erata$^{2}$, Joseph Lilien$^{2}$,}\\[2pt] \textbf{Aman Goel$^{2}$, Ali Torkamani$^{2}$}\\[1em] $^{1}$Columbia University \qquad $^{2}$Amazon Web Services\\[0.8em] \texttt{hantao.yu@columbia.edu}\\ \texttt{\{xxhan,ughai,erata,lilienj,goelaman,alitor\}@amazon.com} }}

\usepackage[most]{tcolorbox}
\definecolor{promptblue}{HTML}{DBEAFE}

\definecolor{posgreen}{HTML}{2F9E44}
\definecolor{negred}{HTML}{E03131}
\iclrfinalcopy
\begin{document}

\maketitle
\fancyhead{}                       % Remove the publication header

\begin{abstract}
On-Policy Context Distillation (OPCD) has recently emerged as a powerful technique for transferring context to student models and for self-improvement. In OPCD, the teacher is conditioned on privileged information, and the goal is to minimize the Kullback-Leibler (KL) divergence between the privileged teacher and the student, evaluated on student-generated tokens. Many existing studies show that using instance-specific gold answers or gold demonstrations as the default privilege can hurt training performance, especially out-of-distribution (OOD). In this work, we instead design general instructions that target common student mistakes observed on the training samples, and show that such simple instructions can outperform gold as the OPCD privilege. In autoformalization tasks, using a matched formatting instruction as the privilege could outperform gold in OOD accuracy by a large margin. In 7 out of 8 experiments using ProverQA, ProofWriter, and ProntoQA as datasets, and Qwen3-Thinking and Olmo3-Thinking families as models, matched instruction privileges outperform gold in OOD by 4 to 17 points, while remaining on par with gold in-domain. Each instruction is only a few sentences (and thus contains much less information compared to all instance-specific gold) and is applied uniformly to every training sample. These results indicate that a general instruction, which applies equally to source and target domain examples, can be substantially more transferable than instance-specific gold in OPCD while maintaining in-domain performance.
\end{abstract}

\section{Introduction}

Context Distillation (CD) transfers behaviors from a teacher conditioned on privileged context to a student, and is a common approach to task-specific post-training. Prior work shows that CD can internalize instructions, explanations, concrete demonstrations, output schema and factual knowledge \citep{2022arXiv220915189S, choi-etal-2023-fixed, caccia2025trainingplugnplayknowledgemodules}. CD methods usually minimize the tokenwise Kullback-Leibler (KL) divergence between the teacher and the student conditioning on teacher-generated tokens \citep{2021arXiv211200861A, 2022arXiv220915189S}, i.e. off-policy context distillation. However, off-policy distillation means that the student is only trained on traces that teacher generates most frequently, so students cannot correct mistakes on traces that it puts more weights on, resulting in compound errors. Consequently, On-Policy (Context) Distillation (OPCD) has been a popular technique designed to mitigate this distribution-mismatch issue \citep{ICLR2024_5be69a58, gu2024minillm, 2026arXiv260118734Z, 2026arXiv260212275Y, kim2026selfdistillation, kaur2026rethinking, pan2026rlcsd, kim2026opsdcompressesrlvrteaches, hubotter2026reinforcement}, where KL divergence is minimized conditioning on student-generated tokens.

Recently, studies have shown that adding privilege contexts to the teacher could degrade OPCD training performance, especially OOD performance, across math and science \citep{li2026demopsddisagreementmodulatedpolicyselfdistillation, kaur2026rethinking, jukić2026geometricselfdistillationreasoninggeneralization, wang2026tracedistillingmatterstokenrouted}. However, the degradation observed by these studies use \emph{instance-specific gold demonstration or answer} as teacher privileges, meaning that the privilege is a correct reference solution to \emph{each} problem. Additionally, since gold privileges are often harder to build and could be noisy, other privileges such as system prompts have been used as privilege contexts in OPCD \citep{2026arXiv260212275Y, zhu2026facesonpolicydistillationpitfalls, wang2026contextreturnsrobustinternalization, rezaei2026rubricguidedselfdistillationposttrainingrubric}. However, there is limited understanding of whether domain-specific instructions could obtain better in-domain and OOD performance than instance-specific gold privilege in OPCD. Therefore, we ask the following question:
\begin{center}
    \textit{Can short and reusable instruction privileges improve in-domain student performance while supporting better cross-dataset generalization than instance-specific gold in OPCD?}
\end{center}

Our question makes autoformalization the perfect testbed because it is easy to understand the mistakes the model makes and therefore design targeting privileges. Autoformalization is a crucial area to enable large scale math research and verification thanks to the math breakthroughs and their formalization efforts using frontier models \citep{openai2026discretegeometry, openai2026tenadvances, anthropic2026fermatslasttheorem}.  Concretely, autoformalization requires the model to translate natural language statements (that include several premises and a question based on these premises) to some formal targets that can be solved to a verdict: \textsc{True}, \textsc{False} or \textsc{Unknown} by a solver. The goal is to make solver output equal to the answer to the question. In this paper, we focus on first-order-logic (FOL), which is the standard target for logical reasoning. See Section \ref{sec: autoformalization} for a detailed explanation of autoformalization and privilege design in OPCD, and Section \ref{sec: why autoformalization} for a detailed explanation why we believe autoformalization is the ideal testbed for answering this question.

\begin{figure}[t]
    \centering
    \includegraphics[width=\linewidth]{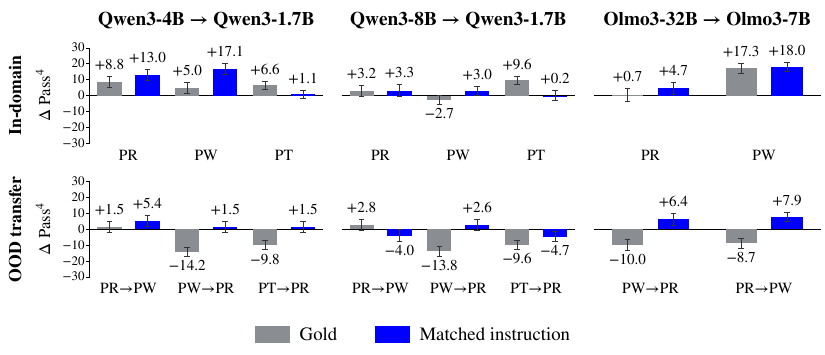}
    \caption{Pass$^4$ accuracy change of using gold FOL and one matched instruction for each dataset as privileges, compared to not using any privilege baseline (OPD). PR, PW, PT refer to ProverQA, ProofWriter and ProntoQA datasets. Model $A \rightarrow$ Model $B$ refers to teacher-student pairs in OPCD, and dataset $X \rightarrow$ dataset $Y$ refers to OOD performance trained on dataset $X$ and evaluated on dataset $Y$. The error bars denote $\pm 1$ standard error.}
    \label{fig:placeholder}
\end{figure}

\textbf{Main Results.} Using Qwen3-Thinking family \citep{qwen3} and Olmo3-Thinking family \citep{olmo2026olmo3} as our models and standard logical datasets including ProverQA \citep{qi2025large}, ProofWriter \citep{tafjord-etal-2021-proofwriter} and ProntoQA \citep{PrOntoQA}, we give a positive answer to our main question (see Figure \ref{fig:placeholder} as a highlight of experimental results).

\textbf{Gold privilege helps in-domain but hurts OOD.} Our experiments indicate that using gold FOL as privilege generally improves student's in-domain performance but hurts OOD performance, aligning with existing OPCD results in math and science.

\textbf{Instruction privilege helps in-domain while avoiding most OOD degradation.} We identify formatting problems the student makes on training sets, and design generic formatting instructions as teacher privileges. Shown in Figure \ref{fig:placeholder}, our instructions outperform gold in most OOD settings (7 out of 8) and have similar performances compared to gold in-domain (5 out of 8) \footnote{We give explanations for why gold dominates in-domain ProntoQA in Section \ref{sec: experimental results}: ProntoQA is cleaner in formatting so formatting instructions are less useful.}. Our instructions are simply a few sentences that only contain a specific formatting instruction (e.g. ``translating XOR relationship to $\oplus$"). Compared to gold privileges that need to contain the correct FOL formulas for each problem, our instructions contain much less information, and are applied uniformly to every single problem. These improvements indicate that simple instructions targeting source and target datasets can be well internalized by OPCD, and we can use much less information in privileges to obtain better training results.

\section{Preliminaries}

In this section, we will explain the basics of context distillation and autoformalization. Throughout, we will use $\theta_T$ to denote the teacher distribution, $\theta_S$ to denote the student distribution, $x$ to denote the prompt and $y$ to denote the output. $\KL$ is the KL divergence.

\subsection{Knowledge Distillation for LLMs}

Knowledge distillation \citep{44873, kim-rush-2016-sequence, DBLP:journals/corr/abs-1910-01108} lets the student mimic the teacher's behavior by optimizing towards the teacher's soft probability distribution at any given sequence. In particular, given a base input or prompt $x$ and a reference output sequence $y$, the loss (average per-token KL divergence) is defined as
\[
\mathcal{L}_{\textup{KD}}(\theta_T,\theta_S,x,y) = \frac{1}{|y|}\sum_{i=1}^{|y|} \KL(\theta_T(\cdot|x,y_{<i})||\theta_S(\cdot|x,y_{<i})).
\] In off-policy knowledge distillation, $y$ is generated by the teacher $y \sim \theta_T(\cdot|x)$. However, since $\theta_T$ and $\theta_S$ are different distributions, it is possible that the student encounters partial sequences that are not usually encountered by the teacher, resulting in under-performance as the student has not been trained on such partial sequences. On-policy distillation is designed to navigate this problem: $y \sim \theta_S(\cdot|x)$ is now chosen from student's own distribution such that the teacher can correct the student's mistakes at the student's own traces.

\subsection{Context Distillation for LLMs}
\label{sec: context distillation for LLMs}

Context distillation (CD) provides the teacher with special privilege $c$ that is embedded in the prompt. For example, $c$ could be the gold demonstration or label, and standard knowledge distillation is when $c = \emptyset$. Consequently, the teacher's distribution becomes $\theta_T(\cdot |c,x,y_{<i})$, and we minimize the KL divergence between privileged teacher's distribution and student distribution. For standard off-policy context distillation, the loss thus becomes
\[
\mathcal{L}_{\textup{off}}(\theta_T,\theta_S,x) = \E_{y \sim \theta_T(\cdot|c,x)}\Big[\frac{1}{|y|}\sum_{i=1}^{|y|}\KL(\theta_T(\cdot|c,x,y_{<i})||\theta_S(\cdot|x,y_{<i}))\Big].
\] For OPCD, $y$ is sampled from student distribution, and the loss is (notice that the only difference is the distribution of $y$ under expectation)
\[
\mathcal{L}_{\textup{on}}(\theta_T,\theta_S,x) = \E_{y \sim \theta_S(\cdot|x)}\Big[\frac{1}{|y|}\sum_{i=1}^{|y|}D_{\textup{KL}}(\theta_T(\cdot|c,x,y_{<i})||\theta_S(\cdot|x,y_{<i}))\Big].
\] When the teacher and student are the same model, OPCD becomes On-Policy Self-Distillation (OPSD). 

\subsection{Autoformalization}
\label{sec: autoformalization}

\begin{figure}[t]
\begin{center}
    \includegraphics[width=\linewidth]{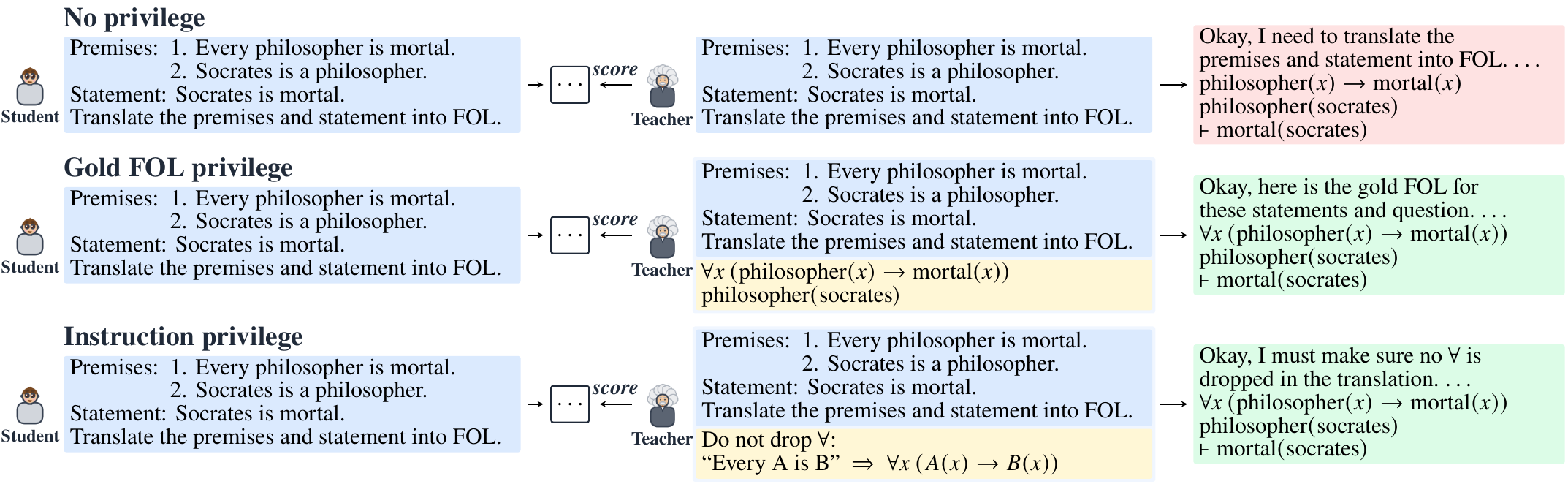}
\end{center}
\caption{Illustration of training students on autoformalization with OPCD. Standard OPD uses no privilege in the teacher, and standard OPCD uses gold FOL. This paper studies using different types of formatting instructions as privileges.}
\label{fig: opcd}
\end{figure}

Autoformalization is the task of translating natural language to some formal targets. In this paper, we focus on first-order logic (FOL), which is the standard target for logical reasoning. Each sample in the dataset contains a set of premises and a question, and the model translates all of them into FOL formulas. The Vampire solver \citep{kovacs2013vampire} takes in the FOL translation and returns the verdict to the question (see Figure \ref{fig: opcd} for an example of training students on autoformalization using OPCD). 

Note that a model can make many mistakes during translation. For example, it might forget to close a parenthesis, translate the same predicate to different names that the solver cannot recognize (\emph{like} and \emph{likes} are lexically identical, but a solver does not know and will identify them as separate predicates), or hallucinate a premise that never exists. Figure \ref{fig: opcd} illustrates the failure mode where $\forall$ is dropped in the translation. All these errors could make the solver output the wrong verdict.

\section{Why Autoformalization is ideal for our goals}
\label{sec: why autoformalization}

Autoformalization is well suited to studying \emph{what makes a good privilege} because it
pairs an objective verifier with a rich, decomposable privilege space. First, correctness is
decided by a symbolic solver (e.g. Vampire, z3) rather than a noisy judge: every rollout yields a discrete
\textsc{True}/\textsc{False}/\textsc{Unknown} verdict. Second, the failure modes are \emph{discrete and diagnosable}: a formula
is unparseable, drops a $\forall$, mistranslates an XOR, or names one concept two ways, so we can attribute each wrong verdict to a specific cause and write a several-line instruction that targets it (see Figure \ref{fig:errortypes} below for examples). In free-form generation (e.g. math and science) these errors are
entangled and hard to isolate. Third, the \emph{same} formatting rules govern all three datasets (and likely to all other logical datasets in autoformalization tasks), so we can construct source$\to$target pairs whose
correct privilege is \emph{known to be shared} and directly measure whether a privilege transfers, exactly the property that separates a reusable instruction from instance-specific gold. Finally, both privilege types arise naturally within the same task (a gold FOL formula per instance
vs.\ a dataset-level formatting rule), enabling a clean, controlled head-to-head under identical training and evaluation.

\section{Experimental Setup}
\label{sec: experimental setup}

Our goal is to evaluate how teacher privileges in OPCD affect training in autoformalization. The training objectives for both can be found at Section \ref{sec: context distillation for LLMs}, and the autoformalization task is introduced in Section \ref{sec: autoformalization}. For comparison, we also conduct experiments for off-policy CD and direct question-answering on the same datasets. In direct question-answering experiments, the privilege is the gold verdict (\textsc{True}/\textsc{False}/\textsc{Unknown}). The system prompts that we use can be found in Appendix \ref{sec: system prompts}.

\textbf{Models.} We experiment with the Qwen3-Thinking family \citep{qwen3} and the Olmo3-Thinking family \citep{olmo2026olmo3}. For Qwen3 family, the student is Qwen3-1.7B and the teacher has two scales: Qwen3-4B and Qwen3-8B. For Olmo3 family, the student is Olmo3-7B and the teacher is Olmo3-32B.

\textbf{Datasets.} For training and evaluation data, we use ProverQA \citep{qi2025large}, ProofWriter \citep{tafjord-etal-2021-proofwriter} and ProntoQA \citep{PrOntoQA} datasets for autoformalization and direct question-answering. These three datasets are all standard benchmarks for logical reasoning. We use the gold FOL formulas provided in ProverQA dataset for autoformalization tasks, deterministic regex converters to generate the gold FOL formulas for ProofWriter and ProntoQA, and gold verdict in all datasets for direct question-answering. 

\textbf{Evaluation metrics.} We use average@8 and pass$^4$ verdict accuracy as our main evaluation metrics in both the autoformalization-solver pipeline and direct question-answering. The model is trained once, and the evaluation on a trained model is performed eight times with different random seeds. Average@8 is the percentage of rollouts that give the correct verdict among all $300 \times 8 = 2400$ rollouts, and pass$^4$ is the average, over all test problems, of $\binom{c}{4}\big/\binom{8}{4}$, where $c \in \{0,1,\dots,8\}$ is the number of correct rollouts among the $8$ for that problem. Pass$^4$ rewards \emph{consistency} and grows steeply with $c$ (e.g.\ $c=8,6,4$ contribute $1.0$, $0.21$, and $0.014$, and $c<4$ contributes $0$), making it a stricter measure of whether a model \emph{reliably}, rather than only occasionally, produces the correct answer. Pass$^k$ has been used extensively to evaluate LLMs and agents \citep{yao2024tau, anonymous2026sequentprover}.

\textbf{Implementation details.} For all datasets we use 1200 training samples and 300 evaluation samples. We set maximum length in training and generation consistently to avoid most truncation effects. For distillation, we use full vocabulary logit distillation with forward KL divergence without clipping. Rollouts and evaluation use temperature 0.7. All experiments are conducted on H100 GPUs with LoRA \citep{hu2022lora}. Full experimental details can be found in Appendix \ref{sec: experimental details}.

\section{Experimental Results}
\label{sec: experimental results}

\subsection{Gold FOL Privilege Improves Verdict Accuracy}

\begin{table}[t]
\caption{Autoformalization results on ProverQA, ProofWriter and
ProntoQA. Numbers in parentheses indicate the change of accuracy when gold FOL is added to the teacher prompt as privilege.}
\label{tab:autoformalization}
\vspace{1ex}
\centering
\small
\setlength{\tabcolsep}{3pt}
\resizebox{\textwidth}{!}{%
\begin{tabular}{l cc cc cc}
\toprule
& \multicolumn{2}{c}{ProverQA}
& \multicolumn{2}{c}{ProofWriter}
& \multicolumn{2}{c}{ProntoQA} \\
\cmidrule(lr){2-3}\cmidrule(lr){4-5}\cmidrule(lr){6-7}
& avg@8 & pass$^4$
& avg@8 & pass$^4$
& avg@8 & pass$^4$ \\
\midrule
Base (no distillation)
& \avgnum{53.3}{} & \avgnum{22.5}{}
& \avgnum{56.7}{} & \avgnum{28.9}{}
& \avgnum{67.0}{} & \avgnum{42.5}{} \\
OPSD (temp=1.0)
& \avgnum{53.4}{} & \avgnum{22.4}{}
& \avgnum{54.7}{} & \avgnum{25.9}{}
& \avgnum{67.2}{} & \avgnum{43.3}{} \\
\midrule
\multicolumn{7}{l}{\textit{Teacher: Qwen3-4B}} \\
CD
& \avgnum{58.7}{} & \avgnum{28.0}{}
& \avgnum{73.5}{} & \avgnum{40.8}{}
& \avgnum{90.1}{} & \avgnum{73.2}{} \\
CD with gold
& \avgnum{\textbf{68.4}}{\textcolor{red}{(+9.7)}}
& \avgnum{\textbf{42.7}}{\textcolor{red}{(+14.7)}}
& \avgnum{81.0}{\textcolor{red}{(+7.5)}}
& \avgnum{51.4}{\textcolor{red}{(+10.6)}}
& \avgnum{90.4}{\textcolor{red}{(+0.3)}}
& \avgnum{71.5}{(-1.7)} \\
OPCD
& \avgnum{62.3}{} & \avgnum{33.2}{}
& \avgnum{79.8}{} & \avgnum{53.2}{}
& \avgnum{93.9}{} & \avgnum{82.5}{} \\
OPCD with gold
& \avgnum{66.0}{\textcolor{red}{(+3.7)}}
& \avgnum{42.0}{\textcolor{red}{(+8.8)}}
& \avgnum{\textbf{83.3}}{\textcolor{red}{(+3.5)}}
& \avgnum{\textbf{58.2}}{\textcolor{red}{(+5.0)}}
& \avgnum{\textbf{96.7}}{\textcolor{red}{(+2.8)}}
& \avgnum{\textbf{89.1}}{\textcolor{red}{(+6.6)}} \\
\midrule
\multicolumn{7}{l}{\textit{Teacher: Qwen3-8B}} \\
CD
& \avgnum{65.5}{} & \avgnum{39.9}{}
& \avgnum{87.7}{} & \avgnum{69.8}{}
& \avgnum{90.9}{} & \avgnum{77.3}{} \\
CD with gold
& \avgnum{68.0}{\textcolor{red}{(+2.5)}}
& \avgnum{41.8}{\textcolor{red}{(+1.9)}}
& \avgnum{87.6}{(-0.1)}
& \avgnum{67.7}{(-2.1)}
& \avgnum{95.1}{\textcolor{red}{(+4.2)}}
& \avgnum{84.2}{\textcolor{red}{(+6.9)}} \\
OPCD
& \avgnum{66.2}{} & \avgnum{43.8}{}
& \avgnum{\textbf{91.0}}{}
& \avgnum{\textbf{76.3}}{}
& \avgnum{91.7}{} & \avgnum{79.4}{} \\
OPCD with gold
& \avgnum{\textbf{69.4}}{\textcolor{red}{(+3.2)}}
& \avgnum{\textbf{47.0}}{\textcolor{red}{(+3.2)}}
& \avgnum{89.0}{(-2.0)}
& \avgnum{73.6}{(-2.7)}
& \avgnum{\textbf{96.3}}{\textcolor{red}{(+4.6)}}
& \avgnum{\textbf{89.0}}{\textcolor{red}{(+9.6)}} \\
\bottomrule
\end{tabular}%
}
\par\vspace{-12pt}
\end{table}

Table \ref{tab:autoformalization} reports the Qwen3 results for autoformalization on ProverQA, ProofWriter and ProntoQA when $c = \emptyset$ and the gold FOL. We evaluate base student (Qwen3-1.7B) as a baseline, and perform self-distillation with no privilege using a different temperature (1.0) to exclude distillation effects. We can see from the table that gold privilege consistently helps with distillation for both CD and OPCD. When teacher is Qwen3-4B, adding gold privilege to the teacher improves the pass$^4$ accuracy by $+8.8,+5.0,+6.6$ across three datasets in OPCD. 

\begin{table}[t]
\caption{Direct question-answering results on three benchmarks. Numbers in parentheses indicate the change of accuracy when gold verdict is added to the teacher prompt as privilege.}
\label{tab:directqa}
\vspace{1ex}
\centering
\small
\setlength{\tabcolsep}{4pt}
\resizebox{\textwidth}{!}{%
\begin{tabular}{l cc cc cc}
\toprule
& \multicolumn{2}{c}{ProverQA}
& \multicolumn{2}{c}{ProofWriter}
& \multicolumn{2}{c}{ProntoQA} \\
\cmidrule(lr){2-3}\cmidrule(lr){4-5}\cmidrule(lr){6-7}
& average@8 & pass$^4$
& average@8 & pass$^4$
& average@8 & pass$^4$ \\
\midrule
\multicolumn{7}{l}{\textit{Qwen3-1.7B, no distillation}} \\
Base
& \avgnum{78.5}{} & \avgnum{64.7}{}
& \avgnum{79.0}{} & \avgnum{69.6}{}
& \avgnum{82.8}{} & \avgnum{68.0}{} \\
\midrule
\multicolumn{7}{l}{\textit{Teacher: Qwen3-4B}} \\
CD
& \avgnum{84.2}{} & \avgnum{70.9}{}
& \avgnum{78.0}{} & \avgnum{63.2}{}
& \avgnum{85.9}{} & \avgnum{67.4}{} \\
CD with verdict
& \avgnum{85.4}{\textcolor{red}{(+1.2)}}
& \avgnum{72.1}{\textcolor{red}{(+1.2)}}
& \avgnum{78.2}{\textcolor{red}{(+0.2)}}
& \avgnum{61.2}{(-2.0)}
& \avgnum{85.3}{(-0.6)}
& \avgnum{65.3}{(-2.1)} \\
OPCD
& \avgnum{85.8}{} & \avgnum{75.3}{}
& \avgnum{81.5}{} & \avgnum{70.6}{}
& \avgnum{\textbf{93.5}}{}
& \avgnum{\textbf{82.1}}{} \\
OPCD with verdict
& \avgnum{\textbf{86.2}}{\textcolor{red}{(+0.4)}}
& \avgnum{\textbf{75.4}}{\textcolor{red}{(+0.1)}}
& \avgnum{\textbf{82.1}}{\textcolor{red}{(+0.6)}}
& \avgnum{\textbf{74.0}}{\textcolor{red}{(+3.4)}}
& \avgnum{\textbf{93.5}}{(+0.0)}
& \avgnum{\textbf{82.1}}{(+0.0)} \\
\midrule
\multicolumn{7}{l}{\textit{Teacher: Qwen3-8B}} \\
CD
& \avgnum{84.7}{} & \avgnum{71.8}{}
& \avgnum{82.8}{} & \avgnum{72.7}{}
& \avgnum{96.6}{} & \avgnum{91.0}{} \\
CD with verdict
& \avgnum{86.0}{\textcolor{red}{(+1.3)}}
& \avgnum{74.1}{\textcolor{red}{(+2.3)}}
& \avgnum{83.4}{\textcolor{red}{(+0.6)}}
& \avgnum{72.5}{(-0.2)}
& \avgnum{97.2}{\textcolor{red}{(+0.6)}}
& \avgnum{91.2}{\textcolor{red}{(+0.2)}} \\
OPCD
& \avgnum{86.8}{}
& \avgnum{\textbf{77.9}}{}
& \avgnum{\textbf{85.5}}{}
& \avgnum{\textbf{78.1}}{}
& \avgnum{\textbf{97.9}}{}
& \avgnum{\textbf{95.3}}{} \\
OPCD with verdict
& \avgnum{\textbf{87.4}}{\textcolor{red}{(+0.6)}}
& \avgnum{76.7}{(-1.2)}
& \avgnum{85.4}{(-0.1)}
& \avgnum{77.1}{(-1.0)}
& \avgnum{97.8}{(-0.1)}
& \avgnum{94.8}{(-0.5)} \\
\bottomrule
\end{tabular}%
}
% \par\vspace{-2pt}
\end{table}

For comparison, Table \ref{tab:directqa} reports CD and OPCD results on direct question-answering tasks on ProverQA, ProofWriter and ProntoQA. Both CD and OPCD still improve direct question-answering capabilities, but verdict privilege has almost no effect on improving the accuracy on all datasets.

Additionally, OPCD dominates CD (best performing method, with or without gold) for both autoformalization and direct question-answering. In ProofWriter autoformalization task, OPCD achieves $58.2$ pass$^4$ accuracy compared to CD's $51.4$ when teacher is Qwen3-4B, and achieves $76.3$ pass$^4$ accuracy compared to CD's $69.8$ when teacher is Qwen3-8B. In ProntoQA autoformalization task, the gap is substantial as well: $89.1$ compared to $73.2$ for Qwen3-4B teacher and $89.0$ compared to $84.2$ for Qwen3-8B teacher. In direct question-answering task, OPCD's lead is even larger.

% \subsection{Pass\^{}4 Metric Favors OPCD}

% In both Table \ref{tab:autoformalization} and \ref{tab:directqa} we observe that the gap between OPCD and CD on pass$^4$ is larger than on average@8. In Table \ref{tab:autoformalization}, the average accuracy shift from CD to its OPCD counterpart is $-0.11$ on average@8 but $+2.13$ on pass$^4$. On ProofWriter it is $+1.69$ on average@8 and $+4.4$ on pass$^4$. OPCD achieving larger pass$^4$ gains suggests training with OPCD gains more consistency.

% \subsection{OPCD Scales Better when Teacher is Larger}

% For self-distillation (teacher is Qwen3-1.7B), Table \ref{tab:autoformalization} and \ref{tab:directqa} show that CD is on-bar with OPCD in both autoformalization and direct question-answering. In autoformalization, CD with gold FOL privilege achieves the best performance on both datasets at both metrics. However, when teacher is larger (Qwen3-4B, 8B and 14B), OPCD consistently outperforms CD with or without gold FOL privilege: when the teacher is Qwen3-4B without gold information, OPCD leads CD by 12.4 in ProofWriter pass$^4$ accuracy. The same phenomenon can be seen in direct question-answering, although the gaps are smaller.

\subsection{Formatting Errors are the Problem}

\begin{figure}[t]
\centering
\includegraphics[width=\linewidth]{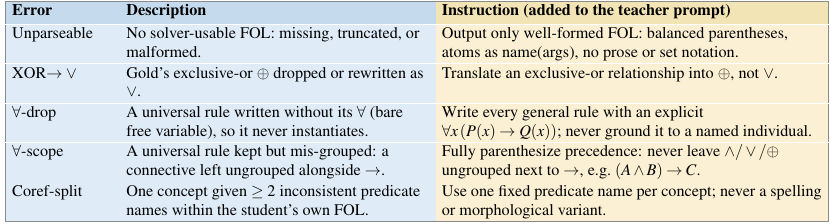}
\caption{Description of all formatting errors and their corresponding instructions. Note that a single rollout could exhibit multiple formatting errors.}
\label{fig:errortypes}
\par\vspace{-15pt}
\end{figure}

We analyze the rollouts that a Qwen3-1.7B model produce on training sets to understand what type of mistakes the student makes. We find out that formatting errors are the major errors, and Figure \ref{fig:errortypes} shows all major formatting error types that we find might be responsible for the wrong verdicts. We observe that many such formatting problems are incorrect translations of connectives and quantifiers, which makes gold FOL a natural candidate to fix them (having a gold FOL demonstration lets the model know how to produce correctly-formatted and parseable FOL formulas). Table \ref{tab:errordecomp-1p7b} reports the error percentages of Qwen3-1.7B on all training sets before any training. It is clear that different datasets have different formatting problems.

\textbf{ProverQA.} Generally suffers from various formatting problems. In particular, XOR$\rightarrow \lor$ (translating XOR relationship to $\lor$) is the most common mistake that Qwen3-1.7B base model makes, while it also has big unparseable, $\forall$-drop and $\forall$-scope problems.
    
\textbf{ProofWriter.} ProofWriter also suffers from various formatting problems which are similar to ProverQA. However, ProofWriter does not have any XOR relationship, so there is no XOR $\rightarrow \lor$ problem in ProofWriter. Dropping $\forall$ is the most common mistake.
    
\textbf{ProntoQA.} The cleanest dataset in terms of formatting problems. Similar to ProofWriter, ProntoQA also does not have any XOR relationship. Dropping $\forall$ and coref-split (translating the same concept into different predicate names so Vampire solver could not identify) are the major problems in terms of formatting. However, ProntoQA suffers the most from semantic problems compared to the other two datasets, meaning that the FOL formulas produced by Qwen3-1.7B model do not have the correct semantic meaning as the natural language.
    
\begin{table}[t]
\caption{Error decomposition for the Qwen3-1.7B student, across all three datasets. The numbers are the percentage of rollouts in the training set that contains a specific type of error. Red denotes the most common problems for each dataset and blue denotes the semantic problem for ProntoQA.}
\label{tab:errordecomp-1p7b}
\vspace{1ex}
\centering
\footnotesize
\setlength{\tabcolsep}{4pt}
\begin{tabular}{l cccccc}
\toprule
Dataset & Unparseable & XOR$\to \lor$ & $\forall$-drop & $\forall$-scope & Coref-split & Semantic \\
\midrule
ProverQA    & 12.5 & \textcolor{red}{30.9} & 14.3 &  7.1 & 2.8 &  2.8 \\
ProofWriter &  8.3 &  0.0 & \textcolor{red}{28.6} & 10.9 & 0.8 &  1.0 \\
ProntoQA    &  4.5 &  0.0 & \textcolor{red}{11.9} &  0.4 & 8.2 & \textcolor{blue}{10.7} \\
\bottomrule
\end{tabular}
\par\vspace{-15pt}
\end{table}

\subsection{Formatting Instructions as Privilege}

We use formatting instructions as privileges in OPCD and compare them with using gold FOL (see Figure \ref{fig:errortypes} for instructions we use). These formatting instructions do not provide any demonstration of what gold FOL should look like for any specific instance, but they inform the teacher specific rules that they should follow when generating the final FOL. Each instruction only contains a single piece of information that is generic to all problems in all datasets, so it contains much less information compared to instance-specific gold FOL formulas. 

As a baseline, we first embed these instructions directly into the prompts to Qwen3-1.7B. Table \ref{tab:direct-prompting} contains the changes when we add the privileges directly into the student prompts. Since gold privilege already contains the answers, adding them into the student prompt gives a substantial improvement overall. However, in-prompt formatting instructions give almost no improvement, and hurt model performance in many cases.

\begin{table}[t]
\caption{Direct-prompting with in-prompt privileges for Qwen3-1.7B (no training): avg@8 and pass$^4$ changes are shown.}
\label{tab:direct-prompting}
\vspace{1ex}
\centering
\small
\setlength{\tabcolsep}{4pt}
\renewcommand{\arraystretch}{1}
\begin{tabular}{@{}l c c c c c c@{}}
\toprule
Dataset & Gold & Unparseable & XOR
& $\forall$-drop & $\forall$-scope & Coref \\
\midrule
ProverQA
& +36.9 / +53.8
& -3.4 / -6.2
& +0.7 / +0.3
& -7.1 / -8.4
& -1.1 / -1.6
& -2.4 / -2.0 \\
ProofWriter
& +35.0 / +51.3
& -3.8 / -5.6
& -1.5 / -3.3
& +8.1 / +0.0
& -4.3 / -6.7
& -4.1 / -7.5 \\
ProntoQA
& +24.5 / +30.2
& -3.7 / -0.8
& -0.2 / +0.9
& -0.3 / -2.7
& -2.5 / +1.1
& -1.8 / -0.1 \\
\bottomrule
\end{tabular}

\end{table}

Table \ref{tab:indomain-p4-opcd} reports the pass$^4$ improvement on ProverQA, ProofWriter and ProntoQA of using these formatting instructions as privileges in OPCD compared to not having any privilege. We can see from table that for ProverQA and ProofWriter, the best instructions dominate gold: Coref-split instruction achieves $+13.0$ and $+3.3$ compared to gold's $+8.8$ and $3.2$ on ProverQA and $\forall$-drop instruction achieves $+17.1$ and $+3.0$ compared to gold's $+5.0$ and $-2.7$ on ProofWriter. On ProntoQA, gold dominates all instructions because as we have seen in Table \ref{tab:errordecomp-1p7b}, ProntoQA is the cleanest dataset with much fewer formatting errors but much higher semantic errors.

\begin{table}[t]
\caption{In-distribution pass$^4$ improvement over no-privilege
(base) OPCD. Teachers are Qwen3-4B and Qwen3-8B and student is
Qwen3-1.7B.}
\label{tab:indomain-p4-opcd}
\vspace{1ex}
\centering
\small
\setlength{\tabcolsep}{5pt}
\renewcommand{\arraystretch}{1}
\begin{tabular}{ll@{\hspace{1.5em}}c@{\hspace{1.5em}}cccccc@{\hspace{1.5em}}c}
\toprule
Dataset & Teacher & +Gold & +Unpars & +XOR
& +$\forall$-drop & +$\forall$-scope & +Coref & Avg & Best \\
\midrule
\multirow{2}{*}{ProverQA}
 & 4B & +8.8 & +9.2 & +5.7 & +0.6 & +5.3
 & \textbf{+13.0} & +6.8 & \textbf{46.2} \\
 & 8B & +3.2 & +2.9 & \textbf{+3.9} & -3.1 & +0.3
 & +3.3 & +1.5 & \textbf{47.7} \\
\midrule
\multirow{2}{*}{ProofWriter}
 & 4B & +5.0 & +5.7 & +1.2 & \textbf{+17.1} & +0.9
 & +0.5 & +5.1 & \textbf{70.3} \\
 & 8B & -2.7 & +2.0 & +2.0 & \textbf{+3.0} & -0.7
 & -4.8 & +0.3 & \textbf{79.3} \\
\midrule
\multirow{2}{*}{ProntoQA}
 & 4B & \textbf{+6.6} & +1.1 & -0.5 & -1.1 & -3.0
 & -15.9 & -3.9 & \textbf{89.1} \\
 & 8B & \textbf{+9.6} & +0.2 & -1.5 & -1.8 & -3.6
 & -4.1 & -2.2 & \textbf{89.0} \\
\bottomrule
\end{tabular}
\par\vspace{-12pt}
\end{table}

We also look at the error percentage change after training (Qwen3-4B teacher) in Table \ref{tab:errordecomp-delta}. Gold privilege generally reduces formatting errors in eval samples, but surprisingly increases semantic problems ($+3.7,+3.2$ for ProverQA and ProofWriter), and some formatting errors ($+12.6$ for Coref in ProverQA). Our instructions avoid these problems in general. A complete table including Qwen3-8B teacher can be found in Appendix \ref{sec: complete error table}.

\begin{table}[t]
\caption{Change (percentage points) in the error decomposition after OPCD with the Qwen3-4B teacher, relative to the no-privilege Qwen3-1.7B student (Table~\ref{tab:errordecomp-1p7b}).}
\label{tab:errordecomp-delta}
\vspace{1ex}
\centering\footnotesize\setlength{\tabcolsep}{4pt}\renewcommand{\arraystretch}{1}
\begin{tabular}{@{}l l@{\hspace{1.5em}}c@{\hspace{1.5em}}cccccc@{}}
\toprule
Data & Priv. & Acc\,$\uparrow$ & Unp.\,$\downarrow$ & XOR\,$\downarrow$ & $\forall$drop\,$\downarrow$ & $\forall$scp\,$\downarrow$ & Coref\,$\downarrow$ & Sem\,$\downarrow$ \\
\midrule
\multirow{3}{*}{ProverQA} & None & +9.0 & -4.5 & -4.9 & -5.8 & +4.0 & +1.0 & -0.5 \\
 & +Gold & +12.7 & -7.0 & -20.6 & -11.4 & +2.5 & +12.6 & +3.7 \\
 & +Coref & \textbf{+14.1} & -8.1 & -5.5 & -5.1 & +3.7 & +1.4 & -1.0 \\
\midrule
\multirow{3}{*}{ProofWriter} & None & +23.1 & -6.5 & 0.0 & -12.4 & -3.7 & -0.6 & -0.8 \\
 & +Gold & +26.6 & -2.4 & 0.0 & -25.7 & -6.0 & -0.5 & +3.2 \\
 & +$\forall$drop & \textbf{+32.1} & -6.6 & 0.0 & -21.9 & -2.9 & -0.6 & -0.8 \\
\midrule
\multirow{3}{*}{ProntoQA} & None & +26.9 & -3.5 & 0.0 & -9.8 & -0.4 & -7.8 & -8.0 \\
 & +Gold & \textbf{+29.8} & -4.0 & 0.0 & -9.9 & -0.4 & -7.9 & -10.2 \\
 & +Unp. & +27.8 & -3.3 & 0.0 & -10.7 & -0.4 & -7.7 & -8.3 \\
\bottomrule
\end{tabular}
\par\vspace{-12pt}
\end{table}

To prove the generalizability of our instructions, we use the best performing instruction on ProverQA (coref-split) and ProofWriter ($\forall$-drop) and compare them against gold in Olmo3 models, using Olmo3-7B thinking as the student and Olmo3-32B thinking as the teacher. Table \ref{tab:olmo-crossfamily-grid} reports the Olmo3 results. For both ProverQA and ProofWriter, matching instruction obtains a slightly better performance compared to gold in-domain ($+4.7$ on ProverQA and $+18.0$ on ProofWriter compared to gold's $+0.7$ and $+17.3$). 

\begin{table}[t]
\caption{In-distribution and OOD pass$^4$ improvement over no privilege OPCD. Teacher is Olmo-32B-Thinking and student is Olmo-7B-Thinking.}
\label{tab:olmo-crossfamily-grid}
\vspace{1ex}
\centering\small
\setlength{\tabcolsep}{5pt}\renewcommand{\arraystretch}{1}
\begin{tabular}{l l c c c c c c}
\toprule
\multicolumn{2}{l}{} & \multicolumn{2}{c}{ProverQA} & \multicolumn{2}{c}{ProofWriter} & \multicolumn{2}{c}{ProntoQA} \\
\cmidrule(lr){3-4} \cmidrule(lr){5-6} \cmidrule(lr){7-8}
Trained on & Privilege & avg@8 & pass$^4$ & avg@8 & pass$^4$ & avg@8 & pass$^4$ \\
\midrule
\multirow{2}{*}{ProverQA}
 & +Gold           & +2.3 & +0.7 & -3.2 & -8.7 & -12.6 & -12.5 \\
 & +Coref          & \textbf{+4.2} & \textbf{+4.7} & \textbf{+2.8} & \textbf{+7.9} & \textbf{-0.9} & \textbf{-1.3} \\
\midrule
\multirow{2}{*}{ProofWriter}
 & +Gold           & -4.6 & -10.0 & +6.6 & +17.3 & -0.6 & -1.0 \\
 & +$\forall$-drop & \textbf{+3.4} & \textbf{+6.4} & \textbf{+7.2} & \textbf{+18.0} & \textbf{+3.7} & \textbf{+5.9} \\
\bottomrule
\end{tabular}
\par\vspace{-12pt}
\end{table}

\subsection{Out-Of-Domain Formatting Instruction Transferring}

Table \ref{tab:ood-p4-opcd} reports the out-of-domain performance of using formatting instructions as privilege during training. Gold does not achieve the best OOD performance in every single row, and there is always a formatting instruction that outperforms gold. In 5 out of 6 cases (except ProverQA$\rightarrow$ProofWriter), gold degrades model performance by a large margin; on the other hand, even the average instructions do not suffer from such degradation. 

We have seen in Table \ref{tab:indomain-p4-opcd} that Coref-split instruction is the best for in-domain ProverQA. For ProofWriter$\rightarrow$ProverQA (ProntoQA$\rightarrow$ProverQA), coref-split instruction obtains $+4.4(+3.3)$ and $-0.9(+4.2)$ respectively and it is still the best instruction overall for OOD on ProverQA. For ProofWriter, $\forall$-drop instruction exhibits a clear win in Table \ref{tab:indomain-p4-opcd}, and it is also the clear winner here for ProverQA$\rightarrow$ProofWriter (ProntoQA$\rightarrow$ProofWriter) as it obtains $+14.5(+7.2)$ and $+8.1(+0.2)$ respectively. Finally, gold privilege performs the best on ProntoQA in-domain but it degrades the model performance OOD: for Qwen3-4B (Qwen3-8B) teacher, $-5.5(-7.6)$ for ProverQA$\rightarrow$ProntoQA and $-6.2(-8.9)$ for ProofWriter$\rightarrow$ProntoQA. Unparseable instruction performs the best overall on these two OOD evaluations. 

Finally, Table \ref{tab:olmo-crossfamily-grid} reports OOD performance given the best in-domain performing instruction for ProverQA and ProofWriter. The best performing instruction (coref-split for ProverQA and $\forall$-drop for ProofWriter) outperforms gold in both OOD eval datasets and under both metrics. Similar to Qwen3 runs, instructions do not suffer from performance degradation when trained on one dataset and evaluated on another.

\section{Related Work}

\textbf{Privilege Design in  Context Distillation.} Context Distillation \citep{2021arXiv211200861A, 2022arXiv220915189S, choi-etal-2023-fixed, caccia2025trainingplugnplayknowledgemodules} is an effective way to transfer knowledge from a teacher to a student without changing teacher's weights. There have been many studies that explore context (privilege) design such as experiential knowledge \citep{2026arXiv260212275Y}, query-specific demonstrations \citep{shenfeld2026selfdistillation}, few-shot examples \citep{yang-etal-2024-self} and split contexts \citep{padmanabhan2026updating}. This work uses autoformalization as a testbed to study privilege design, and we design specific formatting instruction privileges to understand whether they are more transferrable compared to instance-specific gold.

\textbf{On-Policy (Self) Distillation.} On-Policy Distillation (OPD) was designed to mitigate the teacher-student distribution mismatch problem in training. Recent work has studied
distillation in the context of policy \citep{2026arXiv260212275Y}, 
sampling efficiency \citep{ge2026understandingonpolicydistillationlens} in OPD, self-distillation \citep{2026arXiv260118734Z, kim2026opsdcompressesrlvrteaches, hubotter2026reinforcement,penaloza2026privileged} and limitations in policy methods \citep{kaur2026rethinking, li2026demopsddisagreementmodulatedpolicyselfdistillation, jukić2026geometricselfdistillationreasoninggeneralization, wang2026tracedistillingmatterstokenrouted}. This work studies how to design privileges in On-Policy (Self) Distillation.

\textbf{Autoformalization.} The autoformalization-solver pipeline has been well studied in the literature \citep{11229117}, and many prior studies have focus on fine-tuning language models to enable better autoformalization \citep{ICLR2025_3e592c57, thatikonda2026improvingsymbolictranslationlanguage,putra-etal-2026-nl2logic,bansal-etal-2025-robustness}, and most existing work uses off-policy methods. This work designs formatting instructions and use them to understand popular fine-tuning algorithms such as off-policy and on-policy distillation.

\section{Discussions}
We conducted an empirical study of privilege design in On-Policy Context Distillation, where we found that using instructions as privileges could improve student performance while avoiding OOD degradation caused by using gold privileges. The instructions we designed are generic and contain much less information than instance-specific gold privilege, suggesting that providing the teacher full answers might be less effective than providing the teacher a transferrable instruction that targets student's vulnerabilities. 

\textbf{Future work.} This work focuses on autoformalization where the model translates natural language into first-order logic. We explain in Section \ref{sec: why autoformalization} why we believe autoformalization is an ideal testbed for understanding the use of privilege, but we believe this question could be studied in other domains as well. The privileges could look very different in other domains, and we hope our study can inform future work that generic instructions targeting student mistakes can boost training performance. One promising direction is to develop a framework to understand the effectiveness of different  privilege types in distillation, and their tradeoffs in terms of both in-domain and OOD performance.

\begin{table}[t]
\caption{Out-of-distribution transfer (OPCD): pass$^4$ improvement over no-privilege (base) distillation. Student is Qwen3-1.7B and teachers are Qwen3-4B and Qwen3-8B. PR, PW, PT represent ProverQA, ProofWriter and ProntoQA respectively.}
\vspace{1ex}
\label{tab:ood-p4-opcd}
\centering\small\setlength{\tabcolsep}{5pt}\renewcommand{\arraystretch}{1}
\setlength{\aboverulesep}{0.4ex}\setlength{\belowrulesep}{0.5ex}
\begin{tabular}{l l@{\hspace{10pt}}c@{\hspace{10pt}}cccccc@{\hspace{10pt}}c}
\toprule
Transfer & Teacher & +Gold & +Unpars & +XOR & +$\forall$-drop & +$\forall$-scope & +Coref & Avg & Best \\
\midrule
\multirow{2}{*}{PR$\to$PW}  & 4B & $+1.5$  & $+7.5$ & $+6.0$ & $\textbf{+14.5}$ & $+7.4$ & $+5.4$ & $+8.2$ & $\textbf{53.2}$ \\
                                           & 8B & $+2.8$  & $+2.0$ & $+1.6$ & $\textbf{+7.2}$ & $-1.0$ & $-4.0$ & $+1.2$ & $\textbf{60.9}$ \\
\midrule
\multirow{2}{*}{PR$\to$PT}     & 4B & $-5.5$  & $+0.9$ & $+1.6$ & $\textbf{+5.2}$ & $+0.6$ & $+0.1$ & $+1.7$ & $\textbf{57.0}$ \\
                                           & 8B & $-7.6$  & $\textbf{+2.9}$ & $-0.6$ & $-2.9$ & $-2.0$ & $-4.5$ & $-1.4$ & $\textbf{60.6}$ \\
\midrule
\multirow{2}{*}{PW$\to$PR}  & 4B & $-14.2$ & $+0.7$ & $-0.5$ & $+1.5$ & $+0.5$ & $\textbf{+4.4}$ & $+1.3$ & $\textbf{37.3}$ \\
                                           & 8B & $-13.8$ & $+2.3$ & $-2.5$ & $+2.6$ & $-0.6$ & $\textbf{+3.3}$ & $+1.0$ & $\textbf{40.4}$ \\
\midrule
\multirow{2}{*}{PW$\to$PT}  & 4B & $-6.2$  & $+3.2$ & $+0.4$ & $\textbf{+3.7}$ & $+0.7$ & $-1.6$ & $+1.3$ & $\textbf{52.0}$ \\
                                           & 8B & $-8.9$  & $\textbf{+3.6}$ & $-0.2$ & $-0.8$ & $+1.2$ & $-2.7$ & $+0.2$ & $\textbf{60.5}$ \\
\midrule
\multirow{2}{*}{PT$\to$PR}     & 4B & $-9.8$  & $\textbf{+1.5}$ & $-3.1$ & $-1.6$ & $-0.8$ & $-0.9$ & $-1.0$ & $\textbf{27.0}$ \\
                                           & 8B & $-9.6$  & $-4.7$ & $-4.5$ & $-0.9$ & $-3.1$ & $\textbf{+4.2}$ & $-1.8$ & $\textbf{29.6}$ \\
\midrule
\multirow{2}{*}{PT$\to$PW}  & 4B & $-14.2$ & $+3.6$ & $-3.1$ & $\textbf{+8.1}$ & $+0.0$ & $-8.5$ & $+0.0$ & $\textbf{41.5}$ \\
                                           & 8B & $-8.3$  & $\textbf{+1.9}$ & $-3.0$ & $+0.2$ & $-4.2$ & $-6.0$ & $-2.2$ & $\textbf{37.2}$ \\
\bottomrule
\end{tabular}
\par\vspace{-12pt}
\end{table}

\newpage

\bibliography{iclr2027_conference}

@inproceedings{
penaloza2026privileged,
title={Privileged Information Distillation for Language Models},
author={Emiliano Penaloza and Dheeraj Vattikonda and Nicolas Gontier and Alexandre Lacoste and Laurent Charlin and Massimo Caccia},
booktitle={Forty-third International Conference on Machine Learning},
year={2026},
url={https://openreview.net/forum?id=ebZcMQImhG}
}

@misc{anthropic2026fermatslasttheorem,
  author       = {{Anthropic}},
  title        = {{Fermat's Last Theorem} in {Lean} 4},
  year         = {2026},
  howpublished = {GitHub repository},
  url          = {https://github.com/anthropics/fermats-last-theorem},
  note         = {Accessed: September 6, 2026}
}

@misc{openai2026tenadvances,
  author = {{OpenAI}},
  title  = {Ten advances in mathematics and theoretical computer science},
  year   = {2026},
  month  = aug,
  url    = {https://openai.com/index/ten-advances-in-mathematics/},
  note   = {Accessed: September 6, 2026}
}

@misc{openai2026discretegeometry,
  author = {{OpenAI}},
  title  = {An {OpenAI} model has disproved a central conjecture in discrete geometry},
  year   = {2026},
  month  = may,
  url    = {https://openai.com/index/model-disproves-discrete-geometry-conjecture/},
  note   = {Accessed: September 6, 2026}
}

@ARTICLE{2021arXiv211200861A,
       author = {{Askell}, Amanda and {Bai}, Yuntao and {Chen}, Anna and {Drain}, Dawn and {Ganguli}, Deep and {Henighan}, Tom and {Jones}, Andy and {Joseph}, Nicholas and {Mann}, Ben and {DasSarma}, Nova and {Elhage}, Nelson and {Hatfield-Dodds}, Zac and {Hernandez}, Danny and {Kernion}, Jackson and {Ndousse}, Kamal and {Olsson}, Catherine and {Amodei}, Dario and {Brown}, Tom and {Clark}, Jack and {McCandlish}, Sam and {Olah}, Chris and {Kaplan}, Jared},
        title = "{A General Language Assistant as a Laboratory for Alignment}",
      journal = {arXiv e-prints},
         year = 2021,
        month = dec,
          eid = {arXiv:2112.00861},
        pages = {arXiv:2112.00861},
          doi = {10.48550/arXiv.2112.00861},
archivePrefix = {arXiv},
       eprint = {2112.00861},
 primaryClass = {cs.CL},
       adsurl = {https://ui.adsabs.harvard.edu/abs/2021arXiv211200861A}
}

@misc{rezaei2026rubricguidedselfdistillationposttrainingrubric,
      title={Rubric-Guided Self-Distillation: Post-Training Without Rubric Verifiers}, 
      author={MohammadHossein Rezaei and Anas Mahmoud and Zihao Wang and Utkarsh Tyagi and Advait Gosai and Razvan-Gabriel Dumitru and Aakash Sabharwal and Bing Liu and Yunzhong He},
      year={2026},
      eprint={2606.12507},
      archivePrefix={arXiv},
      primaryClass={cs.LG},
      url={https://arxiv.org/abs/2606.12507}, 
}

@ARTICLE{2022arXiv220915189S,
       author = {{Snell}, Charlie and {Klein}, Dan and {Zhong}, Ruiqi},
        title = "{Learning by Distilling Context}",
      journal = {arXiv e-prints},
         year = 2022,
        month = sep,
          eid = {arXiv:2209.15189},
        pages = {arXiv:2209.15189},
          doi = {10.48550/arXiv.2209.15189},
archivePrefix = {arXiv},
       eprint = {2209.15189},
 primaryClass = {cs.CL},
       adsurl = {https://ui.adsabs.harvard.edu/abs/2022arXiv220915189S}
}

@ARTICLE{2026arXiv260212275Y,
       author = {{Ye}, Tianzhu and {Dong}, Li and {Wu}, Xun and {Huang}, Shaohan and {Wei}, Furu},
        title = "{On-Policy Context Distillation for Language Models}",
      journal = {arXiv e-prints},
         year = 2026,
        month = feb,
          eid = {arXiv:2602.12275},
        pages = {arXiv:2602.12275},
          doi = {10.48550/arXiv.2602.12275},
archivePrefix = {arXiv},
       eprint = {2602.12275},
 primaryClass = {cs.CL},
       adsurl = {https://ui.adsabs.harvard.edu/abs/2026arXiv260212275Y}
}

@inproceedings{
kaur2026rethinking,
title={Rethinking On-Policy Self-Distillation for Thinking Models},
author={Simran Kaur and Narutatsu Ri and Yinghui He and Liam H Fowl and Sanjeev Arora},
booktitle={ICML 2026 Workshop on Foundations of Deep Generative Models: Understanding Memorization, Generalization, and Reasoning},
year={2026},
url={https://openreview.net/forum?id=VhCJItwQHn}
}

@article{kim2026selfdistillation,
  title={Why Does Self-Distillation (Sometimes) Degrade the Reasoning Capability of LLMs?},
  author={Kim, Jeonghye and Luo, Xufang and Kim, Minbeom and Lee, Sangmook and Kim, Dohyung and Jeon, Jiwon and Li, Dongsheng and Yang, Yuqing},
  journal={arXiv preprint arXiv:2603.24472},
  year={2026},
  url={https://arxiv.org/abs/2603.24472}
}

@ARTICLE{2026arXiv260118734Z,
       author = {{Zhao}, Siyan and {Xie}, Zhihui and {Liu}, Mengchen and {Huang}, Jing and {Pang}, Guan and {Chen}, Feiyu and {Grover}, Aditya},
        title = "{Self-Distilled Reasoner: On-Policy Self-Distillation for Large Language Models}",
      journal = {arXiv e-prints},
         year = 2026,
        month = jan,
          eid = {arXiv:2601.18734},
        pages = {arXiv:2601.18734},
          doi = {10.48550/arXiv.2601.18734},
archivePrefix = {arXiv},
       eprint = {2601.18734},
 primaryClass = {stat.ML},
       adsurl = {https://ui.adsabs.harvard.edu/abs/2026arXiv260118734Z}
}

@inproceedings{ICLR2024_5be69a58,
 author = {Agarwal, Rishabh and Vieillard, Nino and Zhou, Yongchao and Stanczyk, Piotr and Ramos Garea, Sabela and Geist, Matthieu and Bachem, Olivier},
 booktitle = {International Conference on Learning Representations},
 editor = {B. Kim and Y. Yue and S. Chaudhuri and K. Fragkiadaki and M. Khan and Y. Sun},
 pages = {21246--21263},
 title = {On-Policy Distillation of Language Models: Learning from Self-Generated Mistakes},
 url = {https://proceedings.iclr.cc/paper_files/paper/2024/file/5be69a584901a26c521c2b51e40a4c20-Paper-Conference.pdf},
 volume = {2024},
 year = {2024}
}

@inproceedings{
gu2024minillm,
title={Mini{LLM}: Knowledge Distillation of Large Language Models},
author={Yuxian Gu and Li Dong and Furu Wei and Minlie Huang},
booktitle={The Twelfth International Conference on Learning Representations},
year={2024},
url={https://openreview.net/forum?id=5h0qf7IBZZ}
}

@inproceedings{44873,title	= {Distilling the Knowledge in a Neural Network},author	= {Geoffrey Hinton and Oriol Vinyals and Jeffrey Dean},year	= {2015},URL	= {http://arxiv.org/abs/1503.02531},booktitle	= {NIPS Deep Learning and Representation Learning Workshop}}

@inproceedings{kim-rush-2016-sequence,
    title = "Sequence-Level Knowledge Distillation",
    author = "Kim, Yoon  and
      Rush, Alexander M.",
    editor = "Su, Jian  and
      Duh, Kevin  and
      Carreras, Xavier",
    booktitle = "Proceedings of the 2016 Conference on Empirical Methods in Natural Language Processing",
    month = nov,
    year = "2016",
    address = "Austin, Texas",
    publisher = "Association for Computational Linguistics",
    url = "https://aclanthology.org/D16-1139/",
    doi = "10.18653/v1/D16-1139",
    pages = "1317--1327"
}

@article{DBLP:journals/corr/abs-1910-01108,
  author       = {Victor Sanh and
                  Lysandre Debut and
                  Julien Chaumond and
                  Thomas Wolf},
  title        = {DistilBERT, a distilled version of {BERT:} smaller, faster, cheaper
                  and lighter},
  journal      = {CoRR},
  volume       = {abs/1910.01108},
  year         = {2019},
  url          = {http://arxiv.org/abs/1910.01108},
  eprinttype   = {arXiv},
  eprint       = {1910.01108},
  bibsource    = {dblp computer science bibliography, https://dblp.org}
}

@article{qwen3,
    title={Qwen3 Technical Report}, 
    author={An Yang and Anfeng Li and Baosong Yang and Beichen Zhang and Binyuan Hui and Bo Zheng and Bowen Yu and Chang Gao and Chengen Huang and Chenxu Lv and Chujie Zheng and Dayiheng Liu and Fan Zhou and Fei Huang and Feng Hu and Hao Ge and Haoran Wei and Huan Lin and Jialong Tang and Jian Yang and Jianhong Tu and Jianwei Zhang and Jianxin Yang and Jiaxi Yang and Jing Zhou and Jingren Zhou and Junyang Lin and Kai Dang and Keqin Bao and Kexin Yang and Le Yu and Lianghao Deng and Mei Li and Mingfeng Xue and Mingze Li and Pei Zhang and Peng Wang and Qin Zhu and Rui Men and Ruize Gao and Shixuan Liu and Shuang Luo and Tianhao Li and Tianyi Tang and Wenbiao Yin and Xingzhang Ren and Xinyu Wang and Xinyu Zhang and Xuancheng Ren and Yang Fan and Yang Su and Yichang Zhang and Yinger Zhang and Yu Wan and Yuqiong Liu and Zekun Wang and Zeyu Cui and Zhenru Zhang and Zhipeng Zhou and Zihan Qiu},
    journal = {arXiv preprint arXiv:2505.09388},
    year={2025}
}

@inproceedings{
qi2025large,
title={Large Language Models Meet Symbolic Provers for Logical Reasoning Evaluation},
author={Chengwen Qi and Ren Ma and Bowen Li and He Du and Binyuan Hui and Jinwang Wu and Yuanjun Laili and Conghui He},
booktitle={The Thirteenth International Conference on Learning Representations},
year={2025},
url={https://openreview.net/forum?id=C25SgeXWjE}
}

@inproceedings{tafjord-etal-2021-proofwriter,
    title = "{P}roof{W}riter: Generating Implications, Proofs, and Abductive Statements over Natural Language",
    author = "Tafjord, Oyvind  and
      Dalvi, Bhavana  and
      Clark, Peter",
    editor = "Zong, Chengqing  and
      Xia, Fei  and
      Li, Wenjie  and
      Navigli, Roberto",
    booktitle = "Findings of the Association for Computational Linguistics: ACL-IJCNLP 2021",
    month = aug,
    year = "2021",
    address = "Online",
    publisher = "Association for Computational Linguistics",
    url = "https://aclanthology.org/2021.findings-acl.317/",
    doi = "10.18653/v1/2021.findings-acl.317",
    pages = "3621--3634"
}

@inproceedings{
hu2022lora,
title={Lo{RA}: Low-Rank Adaptation of Large Language Models},
author={Edward J Hu and yelong shen and Phillip Wallis and Zeyuan Allen-Zhu and Yuanzhi Li and Shean Wang and Lu Wang and Weizhu Chen},
booktitle={International Conference on Learning Representations},
year={2022},
url={https://openreview.net/forum?id=nZeVKeeFYf9}
}

@inproceedings{
padmanabhan2026updating,
title={Updating Parametric Knowledge with Context Distillation Retains Post-Training Capabilities},
author={Shankar Padmanabhan and Mustafa Omer Gul and Tanya Goyal},
booktitle={Forty-third International Conference on Machine Learning},
year={2026},
url={https://openreview.net/forum?id=OJsGhlTayF}
}

@inproceedings{
shenfeld2026selfdistillation,
title={Self-Distillation Enables Continual Learning},
author={Idan Shenfeld and Mehul Damani and Jonas H{\"u}botter and Pulkit Agrawal},
booktitle={ICLR 2026 Workshop on Lifelong Agents: Learning, Aligning, Evolving},
year={2026},
url={https://openreview.net/forum?id=HlWA3V6iKF}
}

@misc{ge2026understandingonpolicydistillationlens,
      title={Towards Understanding On-Policy Distillation through the Lens of Test-Time Scaling}, 
      author={Xinmu Ge and Zizhuo Zhang and Yu Huang and Jianing Zhu and Lin Yuan and Wanli Gu and Weichang Wu and Weiran Huang and Xiaolu Zhang and Bo Han and Jun Zhou and Jiangchao Yao},
      year={2026},
      eprint={2608.11829},
      archivePrefix={arXiv},
      primaryClass={cs.LG},
      url={https://arxiv.org/abs/2608.11829}, 
}

@misc{thatikonda2026improvingsymbolictranslationlanguage,
title={Improving Symbolic Translation of Language Models for Logical Reasoning}, 
      author={Ramya Keerthy Thatikonda and Jiuzhou Han and Wray Buntine and Ehsan Shareghi},
      year={2026},
      eprint={2601.09446},
      archivePrefix={arXiv},
      primaryClass={cs.CL},
      url={https://arxiv.org/abs/2601.09446}, 
}

@inproceedings{ICLR2025_3e592c57,
 author = {Ryu, Hyun and Kim, Gyeongman and Lee, Hyemin S. and Yang, Eunho},
 booktitle = {International Conference on Learning Representations},
 editor = {Y. Yue and A. Garg and N. Peng and F. Sha and R. Yu},
 pages = {24935--24964},
 title = {Divide and Translate: Compositional First-Order Logic Translation and Verification for Complex Logical Reasoning},
 url = {https://proceedings.iclr.cc/paper_files/paper/2025/file/3e592c571de69a43d7a870ea89c7e33a-Paper-Conference.pdf},
 volume = {2025},
 year = {2025}
}

@INPROCEEDINGS{11229117,
  author={Bui, Tuan and Le, Trong and Thai, Phat and Nguyen, Sang and Hua, Minh and Pham, Ngan and Bui, Thang and Quan, Tho},
  booktitle={2025 International Joint Conference on Neural Networks (IJCNN)}, 
  title={Speaking in Words, Thinking in Logic: A Dual-Process Framework in QA Systems}, 
  year={2025},
  volume={},
  number={},
  pages={1-8},
  doi={10.1109/IJCNN64981.2025.11229117}}

@inproceedings{putra-etal-2026-nl2logic,
    title = "{NL}2{L}ogic: {AST}-Guided Translation of Natural Language into First-Order Logic with Large Language Models",
    author = "Putra, Rizky Ramadhana  and
      Basuki, Raihan Sultan Pasha  and
      Cheng, Yutong  and
      Gao, Peng",
    editor = "Demberg, Vera  and
      Inui, Kentaro  and
      Marquez, Llu{\'i}s",
    booktitle = "Findings of the {A}ssociation for {C}omputational {L}inguistics: {EACL} 2026",
    month = mar,
    year = "2026",
    address = "Rabat, Morocco",
    publisher = "Association for Computational Linguistics",
    url = "https://aclanthology.org/2026.findings-eacl.317/",
    doi = "10.18653/v1/2026.findings-eacl.317",
    pages = "6035--6051",
    ISBN = "979-8-89176-386-9"
}

@inproceedings{bansal-etal-2025-robustness,
    title = "Robustness of Neurosymbolic Reasoners on First-Order Logic Problems",
    author = "Bansal, Hannah  and
      Kurniawan, Kemal  and
      Frermann, Lea",
    editor = "Kummerfeld, Jonathan K.  and
      Joshi, Aditya  and
      Dras, Mark",
    booktitle = "Proceedings of the 23rd Annual Workshop of the Australasian Language Technology Association",
    month = nov,
    year = "2025",
    address = "Sydney, Australia",
    publisher = "Association for Computational Linguistics",
    url = "https://aclanthology.org/2025.alta-main.1/",
    pages = "1--12",
    ISBN = "1834-7037"
}

@inproceedings{
  PrOntoQA,
  title={Language Models Are Greedy Reasoners: A Systematic Formal Analysis of Chain-of-Thought},
  author={Abulhair Saparov and He He},
  booktitle={The Eleventh International Conference on Learning Representations},
  year={2023},
  url={https://openreview.net/forum?id=qFVVBzXxR2V}
}

@article{hubotter2026reinforcement,
  title = {Reinforcement Learning via Self-Distillation},
  author = {Hübotter, Jonas and Lübeck, Frederike and Behric, Lejs Deen and Baumann, Anton and Bagatella, Marco and Marta, Daniel and Hakimi, Ido and Shenfeld, Idan and Kleine Buening, Thomas and Guestrin, Carlos and Krause, Andreas},
  year = {2026},
  journal = {arXiv preprint arXiv:2601.20802},
}

@inproceedings{
anonymous2026sequentprover,
title={Sequent-Prover: Training Agents for Formal, Checkable {SMT}-based Reasoning},
author={Anonymous},
booktitle={Submitted to ACL Rolling Review - May 2026},
year={2026},
url={https://openreview.net/forum?id=DLMqDyHLTu},
note={under review}
}

@misc{wang2026contextreturnsrobustinternalization,
      title={When Context Returns: Toward Robust Internalization in On-Policy Distillation}, 
      author={Xun Wang and Ruishuo Chen and Zhuoran Li and Yu Chen and Longbo Huang},
      year={2026},
      eprint={2606.11627},
      archivePrefix={arXiv},
      primaryClass={cs.LG},
      url={https://arxiv.org/abs/2606.11627}, 
}

@misc{zhu2026facesonpolicydistillationpitfalls,
      title={The Many Faces of On-Policy Distillation: Pitfalls, Mechanisms, and Fixes}, 
      author={Siqi Zhu and Xuyan Ye and Hongyu Lu and Weiye Shi and Ge Liu},
      year={2026},
      eprint={2605.11182},
      archivePrefix={arXiv},
      primaryClass={cs.AI},
      url={https://arxiv.org/abs/2605.11182}, 
}

@misc{wang2026tracedistillingmatterstokenrouted,
      title={TRACE: Distilling Where It Matters via Token-Routed Self On-Policy Alignment}, 
      author={Jiaxuan Wang and Xuan Ouyang and Zhiyu Chen and Yulan Hu and Zheng Pan and Xin Li and Lan-Zhe Guo},
      year={2026},
      eprint={2605.10194},
      archivePrefix={arXiv},
      primaryClass={cs.AI},
      url={https://arxiv.org/abs/2605.10194}, 
}

@misc{jukić2026geometricselfdistillationreasoninggeneralization,
      title={Geometric Self-Distillation for Reasoning Generalization}, 
      author={Josip Jukić and Ivan Titov},
      year={2026},
      eprint={2607.06855},
      archivePrefix={arXiv},
      primaryClass={cs.LG},
      url={https://arxiv.org/abs/2607.06855}, 
}

@misc{li2026demopsddisagreementmodulatedpolicyselfdistillation,
      title={DemoPSD: Disagreement-Modulated Policy Self-Distillation}, 
      author={Yunhe Li and Hao Shi and Wenhao Liu and Mengzhe Ruan and Hanxu Hou and Zhongxiang Dai and Shuang Qiu and Linqi Song},
      year={2026},
      eprint={2607.02502},
      archivePrefix={arXiv},
      primaryClass={cs.LG},
      url={https://arxiv.org/abs/2607.02502}, 
}

@misc{yao2024tau,
      title={$\tau$-bench: A Benchmark for Tool-Agent-User Interaction in Real-World Domains}, 
      author={Shunyu Yao and Noah Shinn and Pedram Razavi and Karthik Narasimhan},
      year={2024},
      eprint={2406.12045},
      archivePrefix={arXiv},
      primaryClass={cs.AI},
      url={https://arxiv.org/abs/2406.12045}, 
}

@misc{olmo2026olmo3,
      title={Olmo 3}, 
      author={Allyson Ettinger and Amanda Bertsch and Bailey Kuehl and David Graham and David Heineman and Dirk Groeneveld and Faeze Brahman and Finbarr Timbers and Hamish Ivison and Jacob Morrison and Jake Poznanski and Kyle Lo and Luca Soldaini and Matt Jordan and Mayee Chen and Michael Noukhovitch and Nathan Lambert and Pete Walsh and Pradeep Dasigi and Robert Berry and Saumya Malik and Saurabh Shah and Scott Geng and Shane Arora and Shashank Gupta and Taira Anderson and Teng Xiao and Tyler Murray and Tyler Romero and Victoria Graf and Akari Asai and Akshita Bhagia and Alexander Wettig and Alisa Liu and Aman Rangapur and Chloe Anastasiades and Costa Huang and Dustin Schwenk and Harsh Trivedi and Ian Magnusson and Jaron Lochner and Jiacheng Liu and Lester James V. Miranda and Maarten Sap and Malia Morgan and Michael Schmitz and Michal Guerquin and Michael Wilson and Regan Huff and Ronan Le Bras and Rui Xin and Rulin Shao and Sam Skjonsberg and Shannon Zejiang Shen and Shuyue Stella Li and Tucker Wilde and Valentina Pyatkin and Will Merrill and Yapei Chang and Yuling Gu and Zhiyuan Zeng and Ashish Sabharwal and Luke Zettlemoyer and Pang Wei Koh and Ali Farhadi and Noah A. Smith and Hannaneh Hajishirzi},
      year={2026},
      eprint={2512.13961},
      archivePrefix={arXiv},
      primaryClass={cs.CL},
      url={https://arxiv.org/abs/2512.13961}, 
}

@misc{caccia2025trainingplugnplayknowledgemodules,
      title={Training Plug-n-Play Knowledge Modules with Deep Context Distillation}, 
      author={Lucas Caccia and Alan Ansell and Edoardo Ponti and Ivan Vulić and Alessandro Sordoni},
      year={2025},
      eprint={2503.08727},
      archivePrefix={arXiv},
      primaryClass={cs.LG},
      url={https://arxiv.org/abs/2503.08727}, 
}

@inproceedings{choi-etal-2023-fixed,
    title = "Fixed Input Parameterization for Efficient Prompting",
    author = "Choi, Eunbi  and
      Jo, Yongrae  and
      Jang, Joel  and
      Jang, Joonwon  and
      Seo, Minjoon",
    editor = "Rogers, Anna  and
      Boyd-Graber, Jordan  and
      Okazaki, Naoaki",
    booktitle = "Findings of the Association for Computational Linguistics: ACL 2023",
    month = jul,
    year = "2023",
    address = "Toronto, Canada",
    publisher = "Association for Computational Linguistics",
    url = "https://aclanthology.org/2023.findings-acl.533/",
    doi = "10.18653/v1/2023.findings-acl.533",
    pages = "8428--8441"
}

@inproceedings{kovacs2013vampire,
    author    = {Kov{\'a}cs, Laura and Voronkov, Andrei},
    title     = {First-Order Theorem Proving and {V}ampire},
    booktitle = {Computer Aided Verification (CAV 2013)},
    editor    = {Sharygina, Natasha and Veith, Helmut},
    series    = {Lecture Notes in Computer Science},
    volume    = {8044},
    pages     = {1--35},
    year      = {2013},
    publisher = {Springer},
    doi       = {10.1007/978-3-642-39799-8_1}
  }

@misc{kim2026opsdcompressesrlvrteaches,
      title={OPSD Compresses What RLVR Teaches: A Post-RL Compaction Stage for Reasoning Models}, 
      author={Jaehoon Kim and Dongha Lee},
      year={2026},
      eprint={2605.06188},
      archivePrefix={arXiv},
      primaryClass={cs.AI},
      url={https://arxiv.org/abs/2605.06188}, 
}

@article{pan2026rlcsd,
  title={RLCSD: Reinforcement Learning with Contrastive On-Policy Self-Distillation},
  author={Pan, Leyi and Tao, Shuchang and Zhai, Yunpeng and Zhang, Lingzhe and Liu, Zhaoyang and Ding, Bolin and Liu, Aiwei and Wen, Lijie},
  journal={arXiv preprint arXiv:2606.11709},
  year={2026}
}

@inproceedings{yang-etal-2024-self,
    title = "Self-Distillation Bridges Distribution Gap in Language Model Fine-Tuning",
    author = "Yang, Zhaorui  and
      Pang, Tianyu  and
      Feng, Haozhe  and
      Wang, Han  and
      Chen, Wei  and
      Zhu, Minfeng  and
      Liu, Qian",
    editor = "Ku, Lun-Wei  and
      Martins, Andre  and
      Srikumar, Vivek",
    booktitle = "Proceedings of the 62nd Annual Meeting of the Association for Computational Linguistics (Volume 1: Long Papers)",
    month = aug,
    year = "2024",
    address = "Bangkok, Thailand",
    publisher = "Association for Computational Linguistics",
    url = "https://aclanthology.org/2024.acl-long.58/",
    doi = "10.18653/v1/2024.acl-long.58",
    pages = "1028--1043"
}
\bibliographystyle{iclr2027_conference}

\appendix
\section{Appendix}

\subsection{System Prompts}
\label{sec: system prompts}
\textbf{Prompts.} Students and the no-privilege teacher receive the prompt shown below: the
ProverQA-style ``Is the following statement true, false, or uncertain?'' preamble is removed, so the model is asked to formalize a bare statement rather than to decide its truth (the solver decides truth). 

\begin{tcolorbox}[colback=promptblue, colframe=blue!55!black, boxrule=0.5pt, arc=2pt,
  left=6pt, right=6pt, top=4pt, bottom=4pt, fonttitle=\bfseries,
  title={Prompt example in ProverQA}]
\small
\textbf{System.} You translate a natural-language logical-reasoning problem into a first-order
logic (FOL) formalization. You are given PREMISES (context) and a STATEMENT (the question's claim).
Produce FOL for each premise and for the statement. Think step by step about the predicates,
entities, quantifiers, and connectives, then output one fenced \texttt{fol} code block: one FOL
formula per premise (one per line), then a final line \texttt{$\vdash$ \textless conclusion-FOL\textgreater}
giving the FOL of the STATEMENT. Use standard FOL syntax:
$\forall\ \exists\ \land\ \lor\ \lnot\ \rightarrow\ \leftrightarrow\ \oplus$, predicates like
\texttt{likes(Alice, Bob)}, constants capitalized. Do NOT decide whether the statement is
true---a solver does that. Just translate faithfully.

\medskip
\textbf{User.} PREMISES: If Paloma is attentive to details, then she understands lighting and visualizes compositions.
If someone understands lighting, then they may not necessarily compose framing well, and vice versa.
Paloma either buys quality equipment or has a creative vision.
If Paloma has a creative vision, then she either understands lighting or can edit images, but not both.
If Paloma travels extensively and has a good camera, then she can become a professional photographer.
Paloma is detail-oriented. [\emph{15 further premises omitted}]\\
\textcolor{orange}{Privileges (e.g. The gold FOL is ...)}\\
STATEMENT to formalize: Paloma travels extensively.\\
Produce the FOL formalization (premise formulas, then \texttt{$\vdash$ conclusion}).
\end{tcolorbox} 

\subsection{Full Experimental Details}
\label{sec: experimental details}

  \begin{table}[H]

    \caption{Hyperparameters.}
    \vspace{2ex}
  \label{app:tab:hp}
  \centering \small
  \setlength{\tabcolsep}{8pt}
  \begin{tabular}{ll}
  \toprule
  \multicolumn{2}{l}{\textit{Optimization}} \\
  Steps / checkpointing        & 400 steps, save every 50 (8 checkpoints) \\
  Effective batch size         & 8 (per-device 1 $\times$ accum 1 $\times$ 8 GPUs) \\
  Learning rate                & $5\times10^{-6}$ \\
  Optimizer / parallelism      & DeepSpeed ZeRO, CPU optimizer offload (\texttt{cpu\_adam}), 8-way \\
  Training data                & all 1200 rows, no correctness filtering (incorrect traces kept) \\
  Seed                         & 1 (single training seed; 8 evaluation samples) \\
  \midrule
  \multicolumn{2}{l}{\textit{LoRA}} \\
  Rank / $\alpha$              & 64 / 128 \\
  Target modules              & all linear (\texttt{q,k,v,o,gate,up,down}) \\
  \midrule
  \multicolumn{2}{l}{\textit{Objective \& decoding}} \\
  Divergence                  & forward KL, $\beta{=}0$, JSD-clipping off \\
  Temperature                 & 0.7 (trace generation, on-policy rollout, and evaluation) \\
  Token budget                & 8192 tokens, 16384 if more than $5\%$ are truncated \\
  Prompt limits               & student $\leq$1536, teacher $\leq$12288 (fits the gold-privileged prompt) \\
  \midrule
  \multicolumn{2}{l}{\textit{Evaluation \& scoring}} \\
  Samples per problem         & 8 (at $T{=}0.7$) \\
  Checkpoint selection        & peak mean in-domain accuracy over the 8 samples; OOD on the peak only \\
  Accuracy                    & avg@8 (mean correctness over the 8 samples) \\
  pass$^4$                    & mean over problems of $\binom{c}{4}/\binom{8}{4}$, $c=\#$correct of 8 \\
  \bottomrule
  \end{tabular}
  \end{table}

\subsection{Complete Error Type Table}
\label{sec: complete error table}

\begin{table}[H]
\caption{Change in the error decomposition after OPCD, relative to the no-privilege Qwen3-1.7B student on evaluation rollouts.}
\label{tab:errordecomp-delta-complete}
\vspace{1ex}
\centering\footnotesize\setlength{\tabcolsep}{3pt}\renewcommand{\arraystretch}{1}
\setlength{\aboverulesep}{0.4ex}\setlength{\belowrulesep}{0.5ex}
\begin{tabular}{@{}l l l@{\hspace{10pt}}c@{\hspace{10pt}}cccccc@{}}
\toprule
Data & Tch. & Priv. & Acc\,$\uparrow$ & Unp.\,$\downarrow$ & XOR\,$\downarrow$ & $\forall$drop\,$\downarrow$ & $\forall$scp\,$\downarrow$ & Coref\,$\downarrow$ & Sem\,$\downarrow$ \\
\midrule
\multirow{6}{*}{ProverQA} & \multirow{3}{*}{4B} & None & +9.0 & -4.5 & -4.9 & -5.8 & +4.0 & +1.0 & -0.5 \\
 &  & +Gold & +12.7 & -7.0 & -20.6 & -11.4 & +2.5 & +12.6 & +3.7 \\
 &  & +Coref & \textbf{+14.1} & -8.1 & -5.5 & -5.1 & +3.7 & +1.4 & -1.0 \\
\cmidrule(l){2-10}
 & \multirow{3}{*}{8B} & None & +12.9 & -8.7 & -6.4 & -7.2 & +2.5 & +2.0 & +0.2 \\
 &  & +Gold & \textbf{+16.2} & -9.7 & -21.5 & -11.2 & -1.6 & +12.0 & +4.7 \\
 &  & +Coref & +14.1 & -7.9 & -8.1 & -6.2 & +0.6 & +2.1 & -0.6 \\
\midrule
\multirow{6}{*}{ProofWriter} & \multirow{3}{*}{4B} & None & +23.1 & -6.5 & 0.0 & -12.4 & -3.7 & -0.6 & -0.8 \\
 &  & +Gold & +26.6 & -2.4 & 0.0 & -25.7 & -6.0 & -0.5 & +3.2 \\
 &  & +$\forall$drop & \textbf{+32.1} & -6.6 & 0.0 & -21.9 & -2.9 & -0.6 & -0.8 \\
\cmidrule(l){2-10}
 & \multirow{3}{*}{8B} & None & +34.4 & -8.0 & 0.0 & -21.3 & -5.0 & -0.7 & -0.9 \\
 &  & +Gold & +32.3 & -5.1 & 0.0 & -25.6 & -9.5 & -0.6 & +2.7 \\
 &  & +$\forall$drop & \textbf{+35.1} & -7.5 & 0.0 & -22.8 & -4.9 & -0.6 & -0.8 \\
\midrule
\multirow{6}{*}{ProntoQA} & \multirow{3}{*}{4B} & None & +26.9 & -3.5 & 0.0 & -9.8 & -0.4 & -7.8 & -8.0 \\
 &  & +Gold & \textbf{+29.8} & -4.0 & 0.0 & -9.9 & -0.4 & -7.9 & -10.2 \\
 &  & +Unp. & +27.8 & -3.3 & 0.0 & -10.7 & -0.4 & -7.7 & -8.3 \\
\cmidrule(l){2-10}
 & \multirow{3}{*}{8B} & None & +24.7 & -3.7 & 0.0 & -9.9 & -0.4 & -7.2 & -6.2 \\
 &  & +Gold & \textbf{+29.3} & -3.4 & 0.0 & -10.7 & -0.4 & -8.0 & -9.5 \\
 &  & +Unp. & +24.7 & -4.0 & 0.0 & -10.2 & -0.4 & -7.2 & -5.5 \\
\bottomrule
\end{tabular}
\end{table}

\end{document}